\documentclass[11pt,a4paper]{article}
\usepackage[hyperref]{acl2021}
\usepackage{times}
\usepackage{latexsym}

\usepackage[hang,flushmargin]{footmisc}
\usepackage{graphicx,graphics}
\usepackage{amsmath}
\usepackage{colortbl, xcolor}
\usepackage{tabu}
\usepackage{float}
\usepackage{caption}
\usepackage{multirow}
\usepackage{booktabs,makecell,tabularx}
\usepackage{comment}
\usepackage{svg}
\aclfinalcopy

\usepackage{microtype}

\title{NT5?! Training T5 to Perform Numerical Reasoning}

\author{Peng-Jian Yang\textsuperscript{$a$}\thanks{\ \hspace{0.3em}Equal contribution},
	Ying Ting Chen\textsuperscript{$a$}\footnotemark[1],
	Yuechan Chen\textsuperscript{$a$}, Daniel Cer\textsuperscript{$a,b$}
	\\
	\texttt{\{lesterpjy, chentim, sonyachan, dcer\}@berkeley.edu}
	\AND
	{\rm\textsuperscript{$a$}University of California}\\Berkeley, CA\And
	{\rm\textsuperscript{$b$}Google Research}\\Mountain View, CA
}

\date{}

\begin{document}
\maketitle
\begin{abstract}
Numerical reasoning over text (NRoT) presents unique challenges that are not well addressed by existing pre-training objectives.  
We explore five sequential training schedules that adapt a pre-trained T5 model for NRoT.
Our final model is adapted from T5, but further pre-trained on three datasets designed to strengthen skills necessary for NRoT and general reading comprehension before being fine-tuned on the Discrete Reasoning over Text (DROP) dataset. The training improves DROP’s adjusted F1 performance (a numeracy-focused score) from 45.90 to 70.83. Our model closes in on GenBERT (72.4), a custom BERT-Base model using the same datasets with significantly more parameters. We show that training the T5 multitasking framework with multiple numerical reasoning datasets of increasing difficulty, good performance on DROP can be achieved without manually engineering partitioned functionality between distributed and symbol modules.

\end{abstract}

\section{Introduction}

Numerical Reasoning over Text (NRoT) is a reading comprehension task that involves producing an answer to numerical question given a short passage as context. Unlike reading comprehension tasks that can be solved by extracting the answer verbatim from the passage, NRoT usually involves using the question to determine the correct mathematical operation(s) while also identifying the correct values from the passage to use.

Research interest in NRoT has grown with the introduction of the Discrete Reasoning Over Paragraphs (DROP) dataset~\cite{Dua2019}. The majority of DROP examples are number questions involving arithmetic, which has motivated complex models that combine symbolic and neural processing modules~\cite{andor2019, ran2019, chen2020}. The best performing DROP model utilize a symbolic arithmetic module in conjunction with a neural network and other techniques such as ensembling.

We demonstrate in this work that manually engineered partitioning of the functionality between distributed and symbol modules is unnecessary for achieving good performance. Rather, the recently introduced Text-to-Text Transfer Transformer (T5)~\cite{Raffel2019} model is able to internalize NRoT without adaptation. We take full advantage of the multitasking ability of T5 to introduce a sequential training pipeline that is low resource, amiable to experimental cycle, and even achieves good performance using smaller scale models.\footnote{All source codes and sample models are available at \href{https://github.com/lesterpjy/numeric-t5}{https://github.com/lesterpjy/numeric-t5}.}

\begin{figure*}[htb!]
	\centering
	\captionsetup{font=small}
	\includegraphics[width=0.95\linewidth]{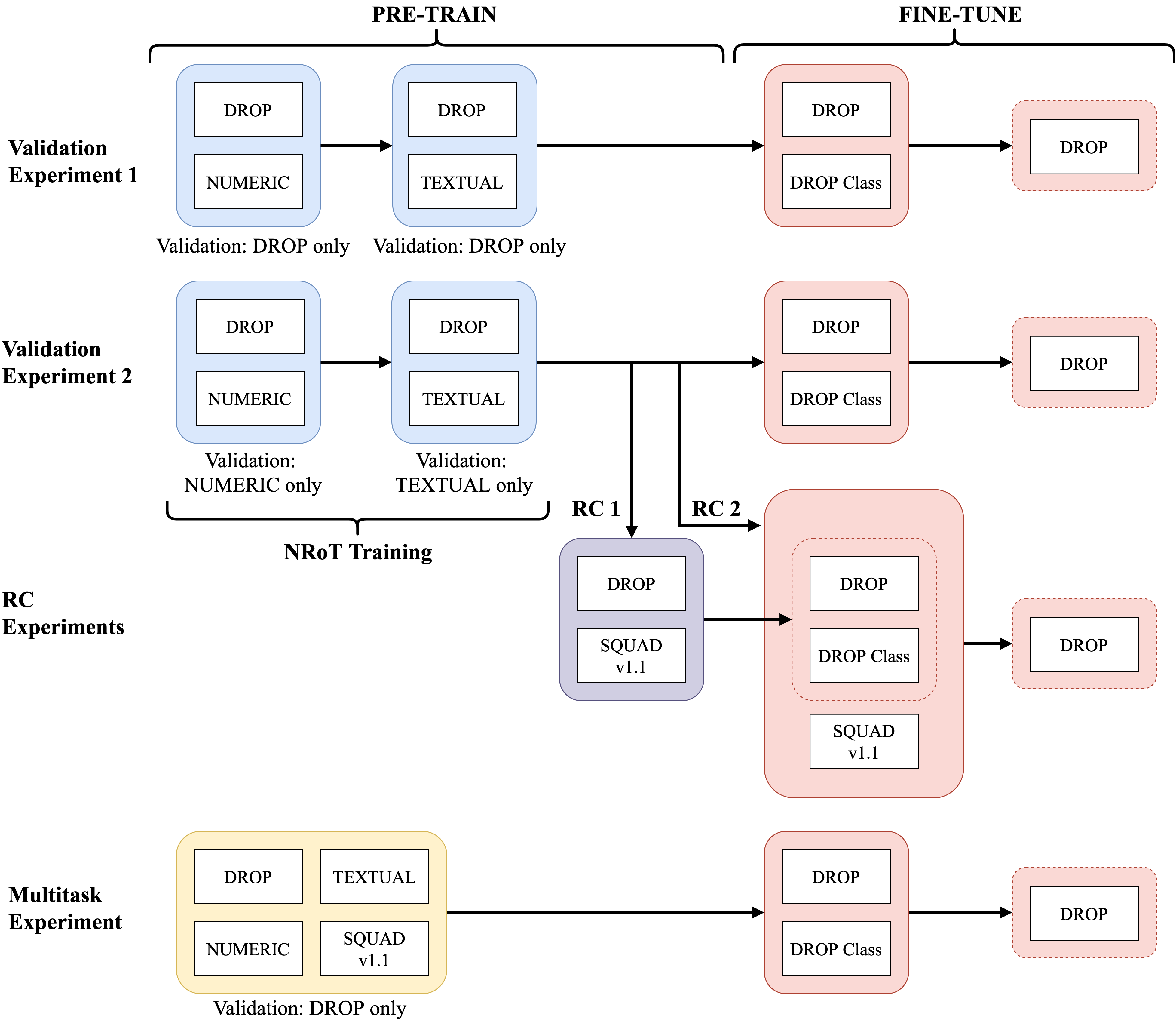}
	\caption{Summary of the five training pipelines. The validation and RC experiments are pre-trained sequentially on NUM, TXT, and SQuAD before fine-tuning on DROP and DROP classification. All experiments begin with pre-trained T5-Small. RC experiment 2 moves the RC training from pre-training to fine-tuning by combining SQuAD, DROP classification, and DROP with multitasking. Multitask experiment is our attempt to multitask-train on all datasets prior to fine-tuning.}
	\label{fig:schedule}
\end{figure*}

\section{T5 for Numerical Reasoning over Text}
We propose five training pipelines for NRoT using T5, each consisting of two stages: pre-training on NRoT and general reading comprehension followed by fine-tuning on DROP and a classification task derived from DROP (Figure \ref{fig:schedule}). Multitask training as described in the T5 paper is used in each stage of training: different datasets are combined with temperature scaling and a special token for identification. Unless specified otherwise, we validate on all the datasets in each respective stage. The first stage begins with a pre-trained T5-Small model~\cite{Raffel2019}. Each following stage, the model begins training on the best performing model from the previous stage.
Our first two configurations (Validation Experiments 1\&2) are designed to test the performance of selecting the best models using different validation data. We experimented with validating on the DROP dev set versus validating on the dev sets of the synthetic datasets (NUM/TXT), described in detail in Section \ref{sec:datasets}.
The next two experiments (RC Experiments 1\&2) attempt to strengthen reading comprehension by multitask training using SQuAD. Finally, we attempt multitasking on all datasets simultaneously (Multitask Experiment). Multitask Experiment is trained with validation on DROP only, instead of validation on the synthetic datasets (NUM/TXT) due to concerns with the model learning parameters for the synthetic datasets closer to fine-tuning and test time. The SQuAD dataset is introduced as an extra step for learning complicated language tasks in RC1 and RC2 for this reason. Since SQuAD is included in the single first step for pre-training in the Multitask experiment, we have deliberately avoided validating its pre-training step on the synthetic datasets. Based on the original T5 paper, we hypothesize that multitasking without stages would be the best way to achieve optimal performance.

\label{sec:pre-exp}

\newcolumntype{e}{X}
\newcolumntype{k}{>{\raggedright\arraybackslash\hsize=.4\hsize}X}
\begin{table*}[ht]
	\centering
	\small
	\captionsetup{font=small}
		\begin{tabularx}{\textwidth}{p{2cm}|e|k|p{1.0cm}}
			\hline
			Reasoning & Passage (shorten) & Question & Answer \\ \hline\hline
			Count + Sort & Denver would retake the lead. . .yet Carolina answered as kicker John Kasay ties the game with a 39-yard field goal. . . . Carolina closed out the half with Kasay nailing a 44-yard field goal. . . . In the fourth quarter, Carolina sealed the win with Kasay’s 42-yard field goal. & Which kicker kicked the most field goals? & John Kasay \\\hline
			
			Subtraction & That year, his Untitled (1981), a painting. . . was sold by Robert Lehrman for 16.3 million, well above its 12 million high estimate. & How many more dollars was the Untitled (1981) painting sold for than the 12 million dollar estimation? & 4300000\\\hline
		\end{tabularx}
	\caption{\label{tbl:drop_ex} Examples of QA pairs found in DROP. The question types and distribution in DROP are subtraction (28.8\%), comparison (18.2\%), selection (19.4\%), addition (11.7\%), count (16.5\%), sort (11.7\%), coreference resolution (3.7\%), other arithmetic (3.2\%), set of spans (6.0\%), other (6.8\%). Combinations of reasoning skills are also possible.}
\end{table*}

\section{Datasets}
\label{sec:datasets}
\paragraph{DROP} Discrete Reasoning Over Paragraphs (DROP), introduced by AllenNLP in 2019~\cite{Dua2019},  is a crowdsourced, adversarially-created 96k question benchmark. The benchmark consists of four types of questions, which can be answered using the context provided. Approximately 61\% of the examples in DROP are number questions that involves arithmetic. The other types are “single-span” (32\%), “spans” (6\%), and “date” (2\%). Note that all four question types in DROP can require NRoT skills, as shown in Table \ref{tbl:drop_ex}. 

\paragraph{Synthetic Data} Two synthetic datasets tailored to boost performance on DROP are developed by~\cite{Geva2020}. The Numeric dataset (NUM) consists of near 1M synthetically generated questions on seven types of numerical skills. Textual dataset (TXT) builds on NUM, and includes 2M plus synthetically generated examples.

We introduce an additional synthetic task based on the DROP dataset itself, whereby the model learns to predict the DROP question-type. While not provided at test time, we expect that explicit awareness of the question types will aid the model in knowing what reasoning strategies to use. 
\paragraph{SQuAD} We investigate using SQuAD v1.1~\cite{rajpurkar2016} to improve NRoT by strengthening general reading comprehension in question and answering tasks. 

\paragraph{Evaluation}
DROP employs two metrics for evaluation: an adjusted F1, and Exact-Match (EM). EM uses that same criteria as SQuAD. F1 has additional logic that invalidates all matching material within an answer when there is a numeric mismatch. Overall F1 is computed using macro-averaging over individual answers. In the presence of multiple ground truths, both EM and F1 will take a max over all computed scores.

\begin{table}[hbt]
	\captionsetup{font=small}
	\small
	\scalebox{0.86}{
		\begin{tabular}{ccccc}
			\hline & \multicolumn{2}{c}{Development} &\multicolumn{2}{c}{Test} \\
			Model & EM & $\text{F}_1$ & EM & $\text{F}_1$ \\ \hline\hline
			Baseline (T5-Small) & 41.12	& 44.64 & 41.97& 45.90 \\
			Validation Experiment 1 & 65.00	& 68.53 & - & - \\
			Validation Experiment 2 & 66.04	& 69.60 & - & - \\
			RC Experiment 1 & \textbf{66.87} & \textbf{70.31} & \textbf{67.00} & \textbf{70.83} \\
			RC Experiment 2 & 66.41	& 69.80 & - & - \\
			Multitask Experiment & 63.10	& 66.47 & - & - \\ \hline\hline
			NAQANet & 46.20 & 49.24 & 44.07 & 47.01 \\
			GenBert & 68.8 & 72.3 & 68.6 & 72.4 \\
			NumNet & 64.92 & 68.31 & 64.56 & 67.97\\
			NumNet+(RoBERTa) & 81.07 & 84.42 & 81.52 & 84.84\\
			QDGAT(RoBERTa) & 84.07 & 87.05 & 84.53 & 87.57\\
			QDGAT(ALBERT) & - & - & 87.04 & 90.10 \\
			
			\hline 
		\end{tabular}
	}
	\caption{\label{tbl:overall} Performance summary for our baseline, training experiments, and select benchmarks. NAQANet is the best-performing model proposed in DROP’s original paper. GenBERT is a modified BERT-base model fine-tuned on the same synthetic datasets. Both NumNet and QDGAT are frameworks with separate language and numerical reasoning modules. QDGAT with an ALBERT language module is the current state-of-the-art.}
\end{table}

\begin{table}[hbt]
	\captionsetup{font=small}
	\small
	\scalebox{1}{
		\begin{tabular}{ccccc}
			\hline
			Model & Initialization &  \#Params\\ \hline
            NT5 & T5-Small & 60M\\
			GenBert & Bert-Base & 110M \\
			NumNet+(RoBERTa) & RoBERTa-Large & 355M\\
			QDGAT(RoBERTa) & RoBERTa-Large & 355M\\
			\hline 
		\end{tabular}
	}
	\caption{\label{tbl:modelsize}Number of parameters used for initialization for respective models.}
\end{table}

\newcolumntype{s}{>{\hsize=.2\hsize}X}
\newcolumntype{g}{>{\hsize=1.6\hsize}X}
\newcolumntype{u}{>{\hsize=.8\hsize}X}
\begin{table*}[hbt]
	\scriptsize
	\centering
	\captionsetup{font=small}
		\begin{tabularx}{\textwidth}{ugssssssssss}
			\toprule & & \multicolumn{2}{X}{Number} & \multicolumn{2}{X}{Date}  & \multicolumn{2}{X}{Span} & \multicolumn{2}{X}{Spans} & \multicolumn{2}{X}{\textbf{Overall}}\\ 
			Model & Schedule & EM & F1 & EM & F1 & EM &F1 & EM &F1 & EM & F1\\ \hline\hline\addlinespace[0.1cm]
			baseline & DROP & 31.79 & 31.83 & 43.95 & 53.28 & 62.09 & 67.42& 26.98 & 55.44 & 41.12 & 44.64 \\
			\addlinespace[0.1cm]\hline\addlinespace[0.1cm]
			
			\multirow{4}{\hsize}{Validation Experiment 1} & DROP + NUM (validate on DROP) & 36.97 & 36.99 & 43.95 & 51.63 & 59.45 & 64.67 & 27.69 & 55.79 & 43.52 & 46.95\\
			\addlinespace[0.1cm]
			& DROP + TXT (validate on DROP) & 63.25 & 63.27 & 42.04 & 51.42 & 63.03 & 68.29 & 29.63 & 56.97 & 60.83 & 64.26\\
			\addlinespace[0.1cm]
			& DROP + DROP class & 67.03 & 67.07 & 42.68 & 51.43 & 65.19 & 70.80 & 31.57 & 58.89 & 63.95 & 67.49\\
			\addlinespace[0.1cm]
			& DROP & 68.72 & 68.78 & 43.31 & 50.36 & 65.09 & 70.62 & 32.10 & 60.05 & 65.00 & 68.53\\
			\addlinespace[0.1cm]\hline\addlinespace[0.1cm]
			
			\multirow{4}{\hsize}{Validation Experiment 2} & DROP + NUM (validate on NUM) & 41.37 & 41.38 & 40.76 & 48.94 & 61.31 & 66.45 & 29.81 & 58.73 & 46.86 & 50.32\\
			\addlinespace[0.1cm]
			& DROP + TXT (validate on TXT) & 63.16 & 63.18 & 44.59 & 52.76 & 63.23 & 68.56 & 29.81 & 58.85 & 60.90 & 64.42\\
			\addlinespace[0.1cm]
			& DROP + DROP class & 67.73 & 67.75 & 45.86 & 53.80 & 65.16 & 70.61 & 33.69 & 61.18 & 64.54 & 68.02\\
			\addlinespace[0.1cm]
			& DROP & 69.78 & 69.83 & 42.68 & 51.23 & 66.00 & 71.47 & 34.22 & 62.53 & 66.04 & 69.60\\
			\addlinespace[0.1cm]\hline\addlinespace[0.1cm]
			
			\multirow{3}{\hsize}{RC Experiment 1}& DROP + SQuAD$^\star$ & 65.61 & 65.68 & 45.22 & 55.38 & 65.94 & 71.23 & 34.39 & 62.75 & 63.52 & 67.06\\
			\addlinespace[0.1cm]
			& DROP + DROP class & 68.65 & 68.69 & 45.22 & 53.87 & 66.54 & 72.01 & 36.68 & 63.47 & 65.71 & 69.17\\
			\addlinespace[0.1cm]
			& DROP &  \textbf{70.34} & \textbf{70.39} & 45.22 & 53.85 & 66.75 & 72.35 &  \textbf{37.74} & 63.43 & \textbf{66.87} & \textbf{70.31}\\
			\addlinespace[0.1cm]\hline\addlinespace[0.1cm]
			
			\multirow{2}{\hsize}{RC Experiment 2}& DROP + DROP class + SQuAD$^{\star\diamond}$ & 65.90 & 65.94 & 45.22 & 54.31 & 66.48 & 71.73 & 35.80 & 62.55 & 63.95 & 67.35\\
			\addlinespace[0.1cm]
			& DROP & 69.44 & 69.47 & 45.22 & 53.17 &  \textbf{67.45} & \textbf{72.60} & 35.63 & 63.08 & 66.41 & 69.80\\
			\addlinespace[0.1cm]\hline\addlinespace[0.1cm]
			
			\multirow{3}{\hsize}{Multitask Experiment} & DROP + TXT + NUM + SQuAD & 56.84 & 56.86 & 42.68 & 50.44 & 64.58 & 69.72 & 33.51 & 61.78 & 57.62 & 61.04\\
			\addlinespace[0.1cm]
			& DROP + DROP class & 63.73 & 63.81 &  \textbf{49.04} & \textbf{56.28} & 65.97 & 71.24 & 36.16 & 63.65 & 62.54 & 65.99\\
			\addlinespace[0.1cm]
			& DROP & 64.43 & 64.49 & 45.86 & 52.99 & 66.48 & 71.61 & 36.51 & \textbf{63.80} & 63.10 & 66.47\\
			\addlinespace[0.1cm]
			
			\bottomrule
		\end{tabularx}
	\caption{\label{tbl:phase-type} The decomposed and overall EM and $\text{F}_1$ scores on different answer types in the development set of DROP for each experiment. High scores for each type are in bold. $\star$Notice that the RC experiments begin training using the weights learned in validation experiment 2. $\diamond$RC Experiment 2 fine-tune with SQuAD in addition to DROP and DROP classification.}
\end{table*}

\section{Results}

The overall results of our five training experiments are summarized in Table \ref{tbl:overall}, and decomposed in Table \ref{tbl:phase-type}.
Our best model achieves 66.8 EM and 70.3 F1 on the dev set, and a 67.0 EM and 70.8 F1 on test. Although the EM and F1 performance appears to have a degree of variance across the experiments. It is clear based on the overall model performance that RC1 and RC2 experiments are the most successful in internalizing the numerical reasoning required for performing well on DROP. While underperforming QDGAT-ALBERT, the current state-of-the-art that makes use of both neural and symbolic modules, NT5 performs well for a purely neural based method. Notably, our models use T5-Small with significantly fewer parameters than GenBERT. The encoder-decoder T5-small model has 60 million parameters, compared to the 110 million parameters of GenBERT in its encoder alone. 

Overall, Table \ref{tbl:phase-type} shows that pre-training with DROP, synthetic datasets and SQuAD, and fine-tuning on DROP and DROP classification sequentially is able to significantly boost the performance on number questions, an increase of F1 from 31.83 to 70.39, while maintaining or improving performance on other types of questions. Additionally, when testing out the baseline, we found T5-Base increase F1 score over T5-Small by 11 points.

\subsection{Difference in Validation Dataset}
A surprising finding here is that saving models while validating on the synthetic dev sets outperforms saving models while validating on the DROP dev sets after the first stage. Specifically, this achieves a F1 score (50.32) that is 3.37 points higher (46.95) without sacrificing performance on span/spans questions. We reason that this performance gap is caused by the difference between the loss on development and DROP’s evaluation metrics, as detailed in Section \ref{sec:datasets}.

\subsection{Strengthening Reading Comprehension}

 Performance on extractive RC tasks is boosted with the addition of SQuAD v1.1 in pre-training. We further test if this performance change persist when multitask training SQuAD v1.1 together with DROP and DROP classification in the fine-tuning stage. The resulting model sees improvement across all question types at the end of the training on SQuAD v1.1. Crucially, performance on RC tasks (date, span, and spans) sees an average improvement of 3.06 points in F1 over the previous result. However, this came at the expense of minor deteriorated performance on numeric questions.

\subsection{All Datasets Multitasking}
Fine-tuning simultaneously on all datasets underperforms our best model by nearly 6-point on F1. 

\section{Error Analysis}
\label{sec:error}
To better understand the achievement and limitations of the best model, we analyzed its errors on the dev set. In 38 of 100 errors sampled from number questions, the model has made at least one partial digit match. Of the total 86 errors on date questions, 39 questions require arithmetic calculations. In 9 of these 86 errors, the model wrongly performs numerical calculations, instead of simply extracting answers. With a sample of 100 span and spans errors, 49 of the questions contain reasoning skills not covered in the pre-training datasets. This is compared to the 43\% shown by~\citet{Dua2019}\footnote{Criteria might vary due to human evaluation}. Many of these errors can be addressed with pre-training datasets that cover more complicated calculations and reasoning skills.

\section{Related Works}
\label{sec:related}

Introduced by~\cite{Geva2020}, GenBERT is a BERT-Base model customized with specialized heads for handling discrete reasoning. It is also the main inspiration for our approach. The current state-of-the-art model on DROP, QDGAT-ALBERT, uses a directed graph attention network between a ALBERT based representation extractor and a prediction module for discrete reasoning~\cite{chen2020}. For works analyzing the mathematical reasoning ability of models over text refer to~\citet{Wallace2019, Ravichander2019}.

\section{Conclusion}
\label{sec:conclusion}

We introduced a sequential pre-training framework for numeracy with T5. Our method demonstrates strong improvements on NRoT over a baseline vanilla T5 model. Although current state of the art, QDGAT, which makes use of a hybrid of a neural and symbol modules, and human performance on DROP are better performing, our approach touts both simplicity and low resource usage, achieving good performance using only T5-small.

\bibliographystyle{acl_natbib}
\bibliography{anthology,acl2021}

\end{document}


\appendix
\section{Appendix}
\label{sec:appendix}

In this appendix we described some of the technical details of our implementation of the NT5 framework, including parameters tuning, and the brief overview of synthetic datasets used. We used Colab Pro TPU as our computing infrastructure, and the average run time for a pipeline including all the steps is roughly 50 hours. 

\subsection{Hyperparameters}
Five input prefixes (context, answer\_me, calculate, and classify\_me) are used for labeling the datasets during training. An example input from DROP is: “answer\_me: How many more men did Ming Rui have than cavalry? context: In March 1768, Ming Rui began his retreat, pursued by a Burmese army of 10,000 men and 2000 cavalry…” Note that we place the question before context in order to avoid the cutoff when the input exceeds our encoder’s max length. Our preset encoder and decoder max lengths (512 and 54, respectively) cover 100\% of context and gold answers in all datasets except for DROP (of which 3.89\% of contexts are cut off in training). Mixing temperatures are set to 1 and training steps are calculated accordingly such that one epoch covers all examples in each of the datasets. Multitask Experiment is the only exception because combining all the datasets in one causes training to extend in time beyond our resource limit. The temperature here is set to 10 and the training steps are calculated such that one epoch is roughly a full run of DROP training set.

\subsection{Learning Rate}
We experiment multiple learning rate schedules on the development set, including constant rates (1e-3, 1e-4, 1e-6) with Adam as the optimizer, and various warm-up/decay rates ([1e-6, 1e-10] and [1e-3, 1e-7] respectively). In the end, we settle on the following parameters: warm-up starting rate (1e-8), warm-up ending rate (1e-4), decay rate after warm-up (1e-3). We allocate 10\% of the total epochs to warm-up, during which the rate increases linearly by batch. Once the warm-up ends, the learning rate decays by epoch according to the following schedule:
\begin{equation}
\text{LR} = \frac{\text{LR}}{(1 + \text{decay rate}(\text{epoch} - \text{warmup epoch}))}
\end{equation}

\subsection{Synthetic Datasets}
\label{appen:syn}
Here we briefly discuss the two synthetic datasets by \citet{Geva2020}. NUM is a dataset consisting of 1M mathematical expressions synthetically generated using six different types of templates that correspond to various numeric operations. An example of the “combination” templates is “s1 f1 s2 f2 s3 f3”, which can translate to a synthetic example of “517.4 - 17484 - 10071.75 + 1013.21.” The symbol “s1” defines the sign for the first digit in the mathematical expression and “f1” defines the first digit to be a float type. Another example is the “max/min/avg” template, “o(f1, f2, f3),” where “o” symbolizes the operation of min/max/avg. The symbols are converted to their associated numbers or operations randomly together with correct answers during data generation. TXT consists of 2M Q\&A examples synthetically generated by building on top of the mathematical skills found in NUM. In addition, the data generation is based on multiple sentence, question, and context templates that are created by following the “world state” framework created by \citet{Hosseini2014} to build math word problems. The extracted templates are sentences with “world states” which consist of countable \textit{entities}, \textit{entity containers} (owners of the entities), and \textit{verbs} (which describe the changes of entities and entity containers) that can be filled. Synthetic context-question pairs with answers can then be randomly generated by filling in the extracted templates with domain-specific vocabularies that correspond to the world states.

\subsection{Temperature-Scaled Mixing}
\label{appen:multitask}
We took advantage of T5 multitasking framework to prevent T5 from unlearning language understanding or numerical reasoning throughout the training schedules. We multitask-train our model by mixing different datasets with temperature-scaled mixing, a strategy utilized in the original work of T5 \citep{Raffel2019} and \textit{How multilingual is Multilingual BERT?}\citep{Pires2019}. This scaling technique is designed to mitigate the issue of a model overfitting to the larger dataset during multitasking due to the magnitude of disparity between datasets. The formula for calculating the temperature scaling rate is as following:

\begin{equation}
r_m = min(length_m \times scale)^{\frac{1}{T}}
\end{equation}

where $T$ is the temperature used across datasets for adjusting mixing proportions, $length_m$ is the size of $dataset_m$, $scale$ modifies the scales of datasets, and $r_m$ is the unnormalized ratio for $dataset_m$. This calculated rate, $r_m$, is then normalized for the sampling ratio of each dataset. When $T = 1.0$, this approach is equivalent to examples-proportional mixing, and no adjustment is done to mitigate the disparity between dataset sizes. The proportions of datasets become closer to equal-mixing as $T$ increases. The greater the $T$ becomes, the more adjustment is done to close the disparity of sizes between datasets, effectively lowering the ratios of larger datasets in the mix.

We decide to set temperature to 1 across all training schedules (except Multiutask Experiment) because we have not found task interference or negative transfer, as described by \citet{Raffel2019}, to be an issue. This is likely due to the fact that the the large datasets (NUM/TXT) are generated by closely following the specific formats and NoRT skills found in DROP.

\bibliographystyle{acl_natbib}
\bibliography{anthology,acl2021}